\makeatletter\def\graphicscache@inhibit{true}\makeatother
\PassOptionsToPackage{hashshortnames}{graphicscache}

\documentclass[letterpaper, 10 pt, conference]{ieeeconf}  %

\IEEEoverridecommandlockouts                              %

\usepackage[utf8]{inputenc}

\usepackage[hidelinks]{hyperref}
\usepackage[style=ieee,hyperref,natbib=true,backend=bibtex,firstinits,doi=false,%
     mincitenames=1,maxcitenames=2,maxbibnames=99,sorting=none,terseinits=false,hyperref=true]{biblatex}
\bibliography{references.bib}
\renewbibmacro*{bbx:savehash}{}%
\defbibheading{bibliography}[\bibname]{\section*{References}}

\usepackage{url}
\usepackage{graphicx}
\usepackage[usenames,dvipsnames]{xcolor}
\usepackage[draft]{fixme}
\fxsetup{theme=color}

\usepackage{amsmath}
\DeclareMathOperator{\atantwo}{atan2}
\usepackage{amssymb}
\usepackage{textcomp}

\usepackage{booktabs}
\usepackage{threeparttable}
\usepackage{multirow}

\usepackage[capitalize]{cleveref}

\usepackage{float}
\usepackage{graphbox}
\usepackage{pgfplots}
\usepackage{pgfplotstable}
\usepgfplotslibrary{groupplots}
\pgfplotsset{compat=newest}
\usepgfplotslibrary{fillbetween}
\usepackage{tikz,tikz-3dplot}
\usetikzlibrary{arrows,arrows.meta,automata,backgrounds,calc,chains,%
decorations.markings,decorations.pathreplacing,decorations.pathmorphing,%
matrix,positioning,shapes,shapes.geometric,shapes.symbols,spy,trees,tikzmark,backgrounds, fit}
\usepackage[binary-units=true,product-units=single,per-mode=symbol,range-units=single,detect-all]{siunitx}
\DeclareSIUnit\pixel{px}

\newcommand{\reffig}[1]{Fig.~\ref{#1}}
\newcommand{\Reffig}[1]{Figure~\ref{#1}}

\newcommand{\refsec}[1]{Sec.~\ref{#1}}

\newcommand{\etal}{et al.~}
\newcommand{\ie}{i.e.,\ }
\newcommand{\eg}{e.g.,\ }

\def\WithAuthorInfo{1}
\usepackage{todonotes}
\usepackage{graphicscache}

\usepackage[export]{adjustbox}

\title{\LARGE \bf
Remote Autonomy for Multiple Small Lowcost UAVs \\ in GNSS-denied Search and Rescue Operations
}

\if\WithAuthorInfo1
\author{Daniel Schleich$^{a}$, Jan Quenzel$^{a,b}$, and Sven Behnke$^{a,b}$%
\thanks{$^{a}$Autonomous Intelligent Systems, University of Bonn, Germany; $^{b}$Center for Robotics and Lamarr Institute for Machine Learning and Artificial Intelligence, University of Bonn, Germany%
}}
\else
\author{Anonymous Authors%
}%
\fi

\begin{document}

\maketitle
\thispagestyle{empty}
\pagestyle{empty}
\begin{tikzpicture}[remember picture,overlay]
\node[anchor=north west,align=left,font=\sffamily, xshift=0.5cm,yshift=-0.5cm] at (current page.north west) {%
	\footnotesize \textbf{Accepted final version.} IEEE International Symposium on Safety, Security, and Rescue Robotics (SSRR), Galway, Ireland, 2025.
};
\end{tikzpicture}%

\begin{abstract}
In recent years, consumer-grade UAVs have been widely adopted by first responders.
In general, they are operated manually, which requires trained pilots, especially in unknown GNSS-denied environments and in the vicinity of structures.
Autonomous flight can facilitate the application of UAVs and reduce operator strain.
However, autonomous systems usually require special programming interfaces, custom sensor setups, and strong onboard computers, which limits a broader deployment.

We present a system for autonomous flight using lightweight consumer-grade DJI drones.
They are controlled by an Android app for state estimation and obstacle avoidance directly running on the UAV's remote control.
Our ground control station enables a single operator to configure and supervise multiple heterogeneous UAVs at once.
Furthermore, it combines the observations of all UAVs into a joint 3D environment model for improved situational awareness.

\end{abstract}

\section{Introduction}
In disaster response scenarios, the ability to quickly obtain an overview of the situation is essential for first responders.
Moreover, the situational picture must be continuously updated throughout the rescue operation.
Unmanned aerial vehicles (UAVs) are increasingly deployed for this task as UAVs cover large inaccessible areas quickly, independent of the terrain.
In most disaster-response operations, trained human pilots directly control the UAVs.
Autonomous flights are commonly restricted to high flight altitudes where the UAV follows preplanned GNSS-waypoint missions without risking collisions~\cite{surmann2022icara}.
Autonomous UAVs that don't need GNSS-based localization~\cite{schleich2021icuas,petrlik2024subt,luo2024sar} rely on custom sensor setups and onboard compute to fly in the vicinity of obstacles.
Thus, larger UAVs with special communication interfaces and sufficient payload are required.
This restricts large-scale deployment and increases risks due to high impact weight and dangerous propellers.

In this work, we propose remote autonomy for GNSS-denied flight of affordable lightweight consumer-grade UAVs, like the DJI Mini\,3\,Pro.
Our solution does not require custom sensors or any hardware modifications, which 
makes it cost-efficient and allows for broad deployment in real-world operations and for using multiple UAV simultaneously to quickly obtain situation-awareness.

In particular, our remote autonomy system includes:
\begin{itemize}
\item an Android app based on the DJI Mobile SDK (MSDK) to control a UAV in GNSS-denied environments, including obstacle avoidance,
\item an environment mapping module that combines the observations of all UAVs into a global model, and
\item an intuitive user interface and fleet management system for configuration and supervision of multiple UAVs.
\end{itemize}

\begin{figure}[t]
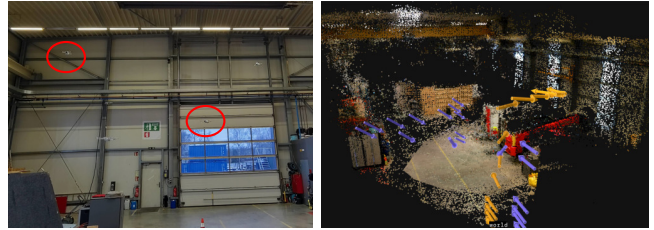

	\centering
    \resizebox{\linewidth}{!}{%
\begin{tikzpicture}
\node[inner sep=0,anchor=north west] (image0) at (0,0)
{\includegraphics[trim=0 0 0 0,clip,height=3cm]{figures/halle_cut_2s.jpg}};

\draw[red, thick] ([xshift=0.6cm,yshift=-1.6cm] image0.north) ellipse [x radius=0.25cm, y radius=0.2cm];

\draw[red, thick] ([xshift=-1.25cm,yshift=-0.75cm] image0.north) ellipse [x radius=0.25cm, y radius=0.2cm];

\node[inner sep=0,anchor=north west,xshift=0.1cm] (image2) at (image0.north east)
{\includegraphics[trim=700 100 175 300,clip,height=3cm]{figures/drz_uav_front_v3.png}};
\end{tikzpicture} 
}
    \vspace{-0.5cm}
	\caption{Multiple consumer-grade UAVs supervised by a single operator facilitate Search \& Rescue in an industrial hall. Left: External view of the UAVs; Right: Created 3D RGB map with UAV trajectories (colored arrows).}
	\label{fig:teaser}
\end{figure}

\section{Related Work}
UAVs are increasingly employed by public safety authorities~\cite{stampa2021jint}, \eg for disaster response~\cite{khan2022jfr,Kruijff-Korbayova:SSRR21}.
This often includes custom hardware designs for the challenges of specific rescue operation, \eg lifebuoy delivery under harsh offshore conditions~\cite{eid2019aset} or avalanche victim search~\cite{silvagni2017avalanche}.
For the task of 3D environment mapping, Lauterbach \etal\cite{lauterbach2019ssrr} equip a large professional-grade UAV with a custom-build sensor unit including LiDAR.
These systems are capable of autonomous flight but rely on GNSS for navigation.
Furthermore, due to the necessary payload, they are usually based on large UAV platforms.

For indoor missions, small UAVs are necessary, which cannot rely on GNSS localization.
One example is the low-cost platform of Tavasoli \etal\cite{tavasoli2023ccee} for damage assessment of structural columns.
The UAV is manually controlled but the movements for data collection can be automated.
Similarly, the solution of Pliakos \etal\cite{pliakos2024JoP} is capable of semi-autonomous GNSS-denied indoor missions, including waypoint navigation, obstacle avoidance and 3D reconstruction of the UAV surroundings.

In contrast, Beul \etal\cite{BeulDNQHB_RAL18} developed a 3D LiDAR-based UAV for SLAM and fully autonomous navigation in warehouses.
Quenzel \etal\cite{QuenzelNDBHB:JINT19} navigate autonomously inside industrial chimneys based on 3D LiDAR and a camera. 
Reyes-Munoz \etal\cite{reyes2021case} demonstrate a system with RGB-D and stereo cameras for fully autonomous flight in GNSS-denied environments.
All of these systems extend commercial air frames with sensors and onboard computers, which increases complexity, weight, and costs; and which decreases their flight time.

For a broader applicability, cost-efficient consumer-grade UAVs are favorable, which do not need hardware modifications.
Following this idea, Khosravi \etal\cite{khosravi2024search} use the DJI Go app and SDK to execute exploration trajectories for autonomous person search on a DJI drone.
However, they consider outdoor environments where the UAV can operate at obstacle-free altitudes using GNSS.

Most similar to our approach is the work of Surmann \etal\cite{surmann2021icara}, who use small consumer-grade UAVs to explore indoor environments.
Their system offers local 3D modeling and autonomous corridor following, while otherwise still relying on manual control.
In contrast, our remote autonomy solution offers fully autonomous flight and globally consistent environment mapping which can incorporate the perception of multiple UAVs.

\section{Remote Autonomy System Setup}
\begin{figure}[t]
  \centering
  \resizebox{0.9\linewidth}{!}{%
\begin{tikzpicture}
[content_node/.append style={font=\sffamily,minimum size=1.5em,minimum width=6em,draw,align=center,rounded corners,scale=0.65},
label_node/.append style={font=\sffamily,scale=0.5},
group_node/.append style={font=\sffamily,dotted,align=center,rounded corners,inner sep=1em,thick},>={Stealth[inset=0pt,length=4pt,angle'=45]}]

\definecolor{yellow}{rgb}  {0.85,0.85,0.0}
\definecolor{red}{rgb}     {1.0,0.0,0.0}
\definecolor{dark_red}{rgb} {0.75,0.0,0.0}
\definecolor{green}{rgb}   {0.0,0.5,0.0}
\definecolor{blue}{rgb}    {0.0,0.0,0.75}
\definecolor{grey}{rgb}    {0.5,0.5,0.5}
\definecolor{redgrey}{rgb}     {0.75,0.5,0.5}

\draw[thick, rounded corners, grey!20!white,fill] (0.0,8) -- (8,8) -- (8,7) -- (0,7) -- cycle;
\draw[thick, rounded corners, grey!20!white,fill] (0.0,6) -- (8,6) -- (8,3) -- (0,3) -- cycle;
\draw[thick, rounded corners, grey!20!white,fill] (0.0,2) -- (8,2) -- (8,0) -- (0,0) -- cycle;

\node(RGB)[content_node,fill=yellow!15!white] at (1.0,7.5) {RGB-Camera};
\node(StateEstimation)[content_node,fill=yellow!15!white] at (3.0,7.5) {State\\Estimation};
\node(ObstacleSensors)[content_node,fill=yellow!15!white] at (5.0,7.5) {Obstacle\\Sensors};
\node(Controller)[content_node,fill=yellow!15!white] at (7.0,7.5) {Flight\\Controller};

\node(VIO_Joint)[dark_red, draw, shape=circle, fill, scale=0.4] at (2.0,6.5) {};

\node(VIO)[content_node,fill=green!15!white] at (1.0,4.5) {VIO};
\node(FSM)[content_node,fill=green!15!white] at (4.0,4.5) {State\\Machine};
\node(ObstacleAvoidance)[content_node,fill=green!15!white] at (7.0,4.5) {Obstacle\\Avoidance};
\node(RCServer)[content_node,fill=green!15!white] at (4.0,3.5) {Protobuf\\Server};

\node(BaseServer)[content_node,fill=blue!15!white] at (4,1.5) {Protobuf\\Server};
\node(UI)[content_node,fill=blue!15!white] at (1.0,0.5) {User\\Interface};
\node(Fleetmanagement)[content_node,fill=blue!15!white] at (4.0,0.5) {Fleet\\Management};
\node(Planner)[content_node,fill=blue!15!white] at (7,0.5) {Planner};
\node(Mapping)[content_node,fill=blue!15!white] at (7,1.5) {Global\\Mapping};

\node(UAV2)[content_node,fill=redgrey!20!white, draw=redgrey!80!white, font=\sffamily] at (1,2.65) {~~RCU \& UAV 2~~};
\node(UAV3)[content_node,fill=redgrey!20!white, draw=redgrey!80!white, font=\sffamily] at (7,2.65) {{UAV 3}};

\node(RC)[content_node, fill=yellow!15!white, minimum width=11.0cm] at (4,5.5) {Remote Controller};
\path[name path=RCtop] (RC.north west) -- (RC.north east);
\path[name path=RCbottom] (RC.south west) -- (RC.south east);

\path[name path=arrow1] (RGB) |- (VIO_Joint) -- ++(0,-1.5) -| (VIO);
\path[name path=arrow2] (ObstacleSensors) -- (5,5.5) -| (ObstacleAvoidance.110);
\path[name path=arrow3] (ObstacleAvoidance) -- (Controller);

\path[name intersections={of=arrow1 and RCtop, by=RCint1_in}];
\path[name intersections={of=arrow1 and RCbottom, by=RCint1_out}];
\draw[->, thick, dark_red, dashed] (RGB) |- (VIO_Joint) -| (RCint1_in);
\draw[-, thick, dark_red, dashed] (StateEstimation)|- (VIO_Joint);
\draw[-, thick, red, opacity=0.5] (RCint1_in) -- (RCint1_out);
\draw[->, thick, red] (RCint1_out) -- ++(0,-0.25) -| (VIO);

\path[name intersections={of=arrow2 and RCtop, by=RCint2_in}];
\path[name intersections={of=arrow2 and RCbottom, by=RCint2_out}];
\draw[->, thick, dark_red, dashed] (ObstacleSensors) -- (RCint2_in);
\draw[-, thick, red, opacity=0.5] (RCint2_in) -- (5,5.5) -| (RCint2_out);
\draw[->, thick, red] (RCint2_out) -- (ObstacleAvoidance.110);

\path[name intersections={of=arrow3 and RCtop, by=RCint3_in}];
\draw[->, thick, dark_red, dashed] (RCint3_in) -- (Controller);
\path[name intersections={of=arrow3 and RCbottom, by=RCint3_out}];
\draw[-, thick, dark_red, dashed, opacity=0.5] (RCint3_in) -- (RCint3_out);
\draw[->, thick, red] (ObstacleAvoidance) -- (RCint3_out);

\draw[->, thick, green] (VIO) -- (FSM);
\draw[<->, thick, green] (RCServer) -- (FSM);
\draw[->, thick, green] (FSM) -- (ObstacleAvoidance);

\draw[<->, thick, dashed, orange] (RCServer) -- (BaseServer);
\draw[<->, thick, dashed, orange] (UAV2) -| (BaseServer.120);
\draw[<->, thick, dashed, black] (UAV3) -| (5.0,1) -| (Fleetmanagement.60);

\draw[<->, thick, blue] (UI) -- (Fleetmanagement);
\draw[<->, thick, blue] (Fleetmanagement) -- (Planner);
\draw[<->, thick, blue] (Fleetmanagement) -- (BaseServer);
\draw[->, thick, blue] (BaseServer) -- (Mapping);
\draw[->, thick, blue] (Mapping) -- (Planner);

\node(UAV_Group_Label)[label_node,anchor=south west] at (-0.3,8.05) {\textbf{\large UAV}};
\node(App_Group_Label)[label_node,anchor=south west] at (-0.3,6.05) {\textbf{\large Remote Control Unit}};
\node(GCS_Group_Label)[label_node,anchor=south west] at (-0.3,2.05) {\textbf{\large Ground Control Station}};

\draw[thin, rounded corners, fill=white] (2,9.05) -- (7.9,9.05) -- (7.9,8.3) -- (2,8.3) -- cycle;
\node[style={font=\sffamily,scale=0.65}] at (0.85,8.8) {};
\draw[->, thick, dashed, dark_red] (2.2,8.8) -- +(0.6,0) node[right,label_node] {Proprietary DJI};
\draw[->, thick, red] (2.2,8.5) -- +(0.6,0) node[right,label_node] {DJI MSDK};
\draw[->, thick, blue] (4.3,8.5) -- +(0.6,0) node[right,label_node] {ROS};
\draw[->, thick, green] (4.3,8.8) -- +(0.6,0) node[right,label_node] {Android App};
\draw[->, thick, dashed, black] (6.3,8.8) -- +(0.6,0) node[right,label_node] {MAVROS};
\draw[->, thick, dashed, orange] (6.3,8.5) -- +(0.6,0) node[right,label_node] {ASIO};

\end{tikzpicture}
}
  \caption{Remote autonomy system architecture. Software modules are divided into two layers: 
  	For each UAV, the safety-relevant modules are executed on the UAV's remote control unit (RCU). 
  	It consists of an Android device and a remote controller (RC), connected via the DJI MSDK. 
  	A WiFi Protobuf server connects each RCU to the ground control station (GCS), where a single operator configures and supervises multiple UAVs and where globally consistent 3D environment models are created and visualized.
  	UAVs from other manufactures can be connected over a MAVROS interface.
  }
  \label{fig:system_architecture}
\end{figure}
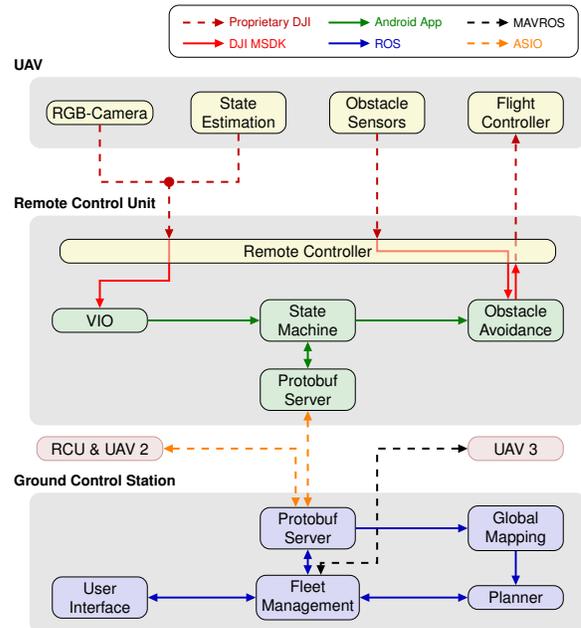

Our remote autonomy system works with a variety of unmodified DJI drones, \eg Mini\,3\,Pro, Mini\,4\,Pro and Mavic\,3T, using the DJI MSDK\footnote{\url{https://github.com/dji-sdk/Mobile-SDK-Android-V5}} (v5.8+).
\Reffig{fig:system_architecture} gives an overview of our approach.

Each UAV connects to its remote controller (RC) via a proprietary wireless link.
The RC connects to an integrated Android device, e.g. in the DJI Pro RC, or an external one, e.g. a smart phone, where the MSDK runs.
To reduce latency as far as possible, all safety-relevant software modules, like visual-inertial odometry (VIO) and obstacle avoidance, run within a single Android app on the RCU.
The app receives new tasks from our ground control station (GCS) and sends images and odometry to the GCS to enable there the fusion of environment measurements from multiple UAVs to a 3D map.
GCS and the RCUs communicate over WiFi using a custom server with Protobuf\footnote{\url{https://github.com/protocolbuffers/protobuf}} for efficient serialization of transmitted messages.

The UAV's h265-encoded video feed is decoded on the RCU for processing within the VIO module.
For efficiency, DJI submits the crucial I-frames exceptionally infrequent.
Missing one I-frame renders the stream useless for one minute or more.
To address this issue, we re-encode the image stream using the RCU's integrated hardware acceleration to ensure correct transmission and conserve bandwidth when forwarding the h265 packages through our Protobuf server.

Our fleet management also has an interface to connect UAVs from other manufacturers over MAVROS\footnote{\url{https://github.com/mavlink/mavros}}.
For these, we assume that the main functionality of our Android app (\eg odometry and obstacle avoidance) is provided by the manufacturer's flight controller.

\subsection{Ground Control Station}
\label{sec:gcs}

\begin{figure*}[!h]
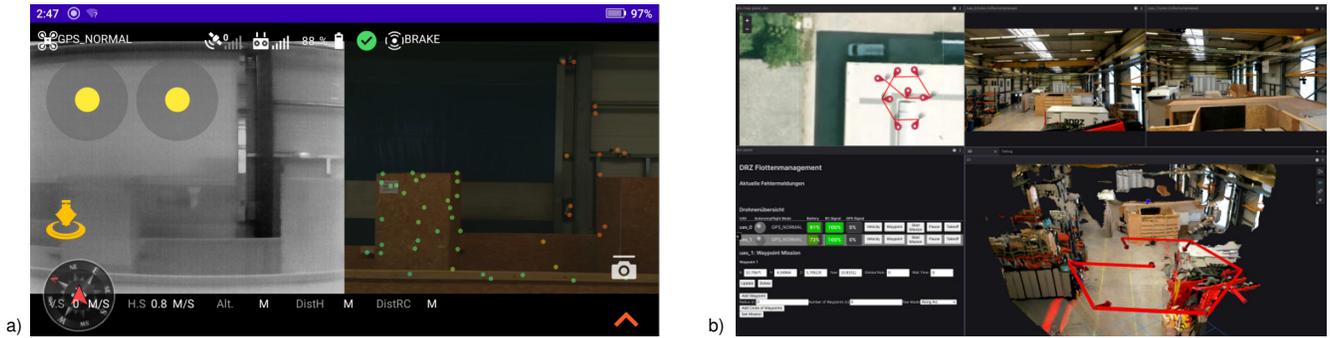

	\centering
	{%
\begin{tikzpicture}
\node[inner sep=0,anchor=north west,] (image0) at (0,0)
{\includegraphics[trim=0 0 0 0,clip,height=4.4cm]{figures/app_ui_in_flight_s.png}};

\node[inner sep=0,anchor=north west,xshift=1cm] (image1) at (image0.north east)
{\includegraphics[trim=0 0 0 0,clip,height=4.4cm]{figures/Interface.png}};

\node[scale=0.75, anchor=south east,xshift=-0.01cm,yshift=-0.1cm, rectangle, fill=white, align=center, font=\small\sffamily] (n_0) at (image0.south west) {a)};

\node[scale=0.75, anchor=south west,xshift=-0.6cm,yshift=-0.1cm, rectangle, fill=white, align=center, font=\small\sffamily] (n_0) at (image1.south west) {b)};

\end{tikzpicture}
} \vspace*{-3ex}
	\caption{User interfaces on our remote controller (RCU, a) and at the ground control station (GCS, b).
		RCU: Various status indicators are overlayed on the video feed for the safety pilot including UAV heading and virtual stick commands.
		The example shows the thermal image and tracked VIO features on the color image of a DJI M3T in side-by-side mode.
		GCS: The fleet management shows the current UAVs' states including control options (b, bottom left).
		The current waypoint mission is shown in the joint 3D environment model (b, bottom right) and overlayed on a satellite image (b, top left). Live-streams (b, top right) allow direct supervision and targeted inspection.
	}
	\label{fig:ui}
\end{figure*}

The GCS consists of a powerful PC for executing all resource-intensive and non-time-critical tasks.
It computes a joint environment model (\refsec{sec:mapping}) from the local perception of each UAV.
Additionally, the GCS visualizes all UAV states in real time and provides an intuitive control interface, allowing a single operator to configure and supervise multiple UAVs simultaneously (\refsec{sec:fleet_management}).
All of these modules are implemented using ROS.

\subsubsection{Globally Consistent 3D Environment Mapping}
\label{sec:mapping}
To generate a global map, we select keyframes from multiple UAVs based on a frustum overlap.
The selection uses our VIO pose estimate as a prior and chooses adjacent keyframe candidates for matching.
We extract XFeat~\cite{potje2024cvpr} from all keyframes and perform matching with LighterGlue between candidates.
Afterwards, we employ sparse incremental reconstruction with the optimization backend of GLOMAP~\cite{pan2024eccv}.

Assuming that multiple UAVs start in the same GNSS-denied area, our mapping initially reconstruct observations from a single UAV.
Another UAV is added once it has enough keyframe matches (\eg 5) with the current reconstruction.
Starting from there, all observations of the UAV will be included.

The optimization backend provides a reconstruction in an arbitrary frame with scale ambiguity.
Our application requires a common metric reference frame, though. 
To address this issue, a RANSAC-based pose alignment facilitates the scale transformation if GNSS data is available.
In GNSS-denied situations, we estimate the scale from the first 10 poses, and align the origin with the first UAV's starting position and orientation.

The resulting map is quite sparse and often difficult to understand quickly, even by trained professionals.
Hence, we establish dense pairwise correspondences using MAST3R~\cite{leroy2024eccv} between our previously matched adjacent keyframes.
Due to time and compute limitations, MAST3R is initially only applied to the new keyframe with two previous ones.
More previously unprocessed pairs are matched by MAST3R in a lazy-evaluation scheme, if no new keyframes are established within some seconds, \eg when all UAVs are hovering.
We merge shared observations from pairwise correspondences between multiple keyframes prior to triangulation.
The triangulation uses the dense matches from MAST3R with our reconstructed metric poses.
\Reffig{fig:teaser} shows estimated camera poses and colored 3D points of autonomous flights of two DJI Mini\,3\,Pro within the industrial hall of the DRZ Living Lab.

\subsubsection{User Interface \& Fleet Management}
\label{sec:fleet_management}

For intuitive supervision and control of our UAVs, we implemented a custom extension to Foxglove Studio\footnote{\url{https://foxglove.dev/studio}} (see \reffig{fig:ui}b).
It continuously monitors the state of all registered UAVs and informs the operator about possible errors.
During missions, new UAVs are added anytime just by connecting the Android app on the corresponding RCU.
The app automatically initiates the connection with the GCS and registers the UAV into the fleet management module.
Here, existing UAVs are recognized on re-establishing the connection after temporary interruption.

In addition to supervision, our UI supports switching between individual UAVs in two control modes.
In \textit{Velocity Control}, the operator steers the UAV using a gamepad similar to the original RCU.
However, directly commanding flight velocities is not feasible due to the latency between control input and video feedback.
Instead, we implemented a \textit{carrot mode} where the operator moves a virtual target pose in the local 3D environment.
This target is visualized without latency and facilitates control significantly.

In \textit{Waypoint Control}, the operator inputs a sequence of target poses as a list of coordinates via keyboard, by moving a marker within the 3D environment model %
using the gamepad, or by marking positions on satellite images or maps.
Additional parameters like individual wait time or gimbal orientation provide flexibility for every mission.
Flight patterns like circles are inserted automatically, without manually defining each waypoint.
To ensure collision-freeness, a dynamic trajectory planner~\cite{schleich2021icra} interpolates between waypoints using the global 3D model.
The fleet management then transforms the resulting flight plan into the local UAV frame and keeps sending the next planned waypoint to the RCU for execution, while monitoring the progress.

\subsection{Remote Controller with Android App}
\label{sec:app}
The RCU runs our Android app (\reffig{fig:ui}a) as a configuration and supervision interface for the safety pilot.
It visualizes the current UAV state, camera live-stream, control mode, and the executed VirtualStick commands.
To enable the autonomy functions, an explicit clearance from the safety pilot is necessary.
Moreover, clearance may be revoked at any time, \eg by moving the control sticks of the RC.

The Android app steers the UAV towards the next target pose using DJI's VirtualStick commands.
These emulate RC stick movements of an operator and represent 3D linear velocities and yaw.
We obtain target velocities from the difference between current and target pose using either GNSS or our state estimation (\refsec{sec:vio}) as position feedback.
These velocities are clipped at a configurable maximum flight speed, and adjusted by the reactive obstacle avoidance (\refsec{sec:obstacle_avoidance}).
Once the UAV is sufficiently close to the target, we switch to DJI's Position Hold mode.
If the UAV drifts too far from the target pose, we reactivate our control.

In addition to the UAV pose, our app controls the gimbal either according to a manually defined orientation from the GCS or automatically during people tracking.
For the latter, we run an EfficientDet Lite0~\cite{tan2020cvpr} at $\approx\SI{20}{\hertz}$ directly on the RCU using the TensorFlow Lite Task library\footnote{\url{https://ai.google.dev/edge/litert/libraries/task_library/overview}}.
Given the person detection, our app adjusts the gimbal to keep the bounding box centered.

\subsubsection{State Estimation}
\label{sec:vio}
Most autonomy functions rely on knowledge of our position and orientation relative to the environment, which is especially important in GNSS-denied environments.
Outdoors, the DJI MSDK provides the UAV's GNSS position with up to $\pm \SI{0.1}{\meter}$ resolution.
In GNSS-denied environments no position information is given, though.
The MSDK's reported velocity and orientation are quantized to an accuracy of $\pm \SI{0.1}{\meter\per\second}$, resp. $\pm\SI{0.1}{\degree}$.
Moreover, current MSDKs (v5.8+) supply the current (quantized) state not on a regular basis, but only once something changes.

As simply integrating the velocity would be too inaccurate and without IMU measurements, we adapted OpenVINS~\cite{geneva2020icra} for the provided inputs:
~i) We merge and retain the current measurements to submit them to the VIO at a required continuous rate.
~ii) We replaced the IMU input during state propagation with a constant velocity (CV) motion model.
~iii) The truncated velocity and orientation are integrated in the update step considering an appropriate uncertainty.
If available, the UAV's altitude is further included in the update.
As a result, our modified VIO runs on the RCU on the monocular FPV camera stream with up to \SI{60}{\hertz}.

\subsubsection{Obstacle Avoidance}
\label{sec:obstacle_avoidance}
In order to react quickly to dynamic or unseen obstacles, we reactively adjust the velocity commands similar to the Predictive Angular Potential Fields method~\cite{schleich2022iros}.
The MSDK provides \num{360} horizontal obstacle measurements in a scan line where only a UAV-depending subset is valid.
A Mini\,3\,Pro supplies valid distances in forward and backward direction, while a Mini\,4\,Pro covers the full \SI{360}{\degree}.
The received distances are aggregated over a short window into a spherical range image.
Before computing the global potential field $\mathcal P$ using a $L_1$ distance transform~\cite{felzenszwalb2012toc}, we define per pixel the force as $\atantwo(d_\mathrm{s}, r)$, with safety distance $d_\mathrm{s}$ and obstacle distance $r$.
To determine the direction, \ie yaw, of the adjusted velocity, we first project the target command into the potential field image.
We then find the closest zero-force pixel within the FoV using gradient descent and reproject it into 3D.
The magnitude of the velocity command is determined by the obstacle distance in the corresponding direction.

In general, we try to maximize the distance to obstacles.
However, close distances cannot be avoided in certain situations, \ie when passing through narrow corridors.
We address this using two different safety distances $d_\text{s}$ and $d_\text{min}$ along with their corresponding potential fields $\mathcal P_\text{s}$ and $\mathcal P_\text{min}$.
If no zero-force pixel is found for $\mathcal P_\text{s}$, we choose a zero-force pixel from $\mathcal P_\text{min}$ minimizing $\mathcal P_\text{s}$.
If such a pixel is not found either, we stop as there is no trajectory with sufficient obstacle clearance.

\subsubsection{Door Traversal Mode}
\label{sec:door_mode}
The measurement range of DJI obstacle sensors has a lower limit of \SI{0.5}{\meter}.
During experiments, we found that closer obstacles are still frequently detected but the distance measurements become unreliable, overestimating the actual values.
Thus, we enforce an obstacle clearance of at least \SI{0.5}{\meter} within our obstacle avoidance.
However, when passing through doors, this clearance cannot be maintained.
Thus, we integrated a specific control mode for autonomous door traversal.

\begin{figure}[t]
	\centering
	\resizebox{0.7\linewidth}{!}{\includegraphics{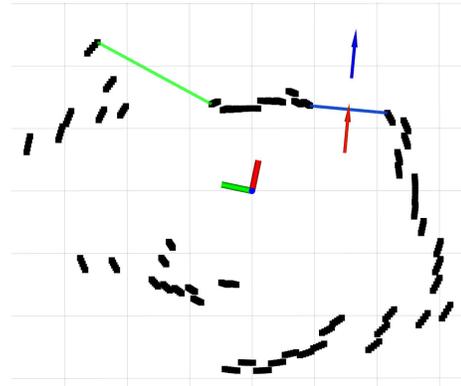}}\vspace*{-1ex}
	\caption{Door detection within horizontal obstacle scan. The current UAV pose is marked by the coordinate axes. The green line indicates a door candidate that is filtered out since its normal deviates too much from the scan's radial direction. The blue line represents the selected door candidate. The red and blue arrows correspond to the planned pre- and post-traversal poses, respectively.}
	\label{fig:door_detection}
\end{figure}

After approaching the door using the control pipeline described above, the operator manually triggers the door traversal mode.
In a first step, the door opening is detected within the horizontal obstacle scan by searching for jumps in adjacent distance measurements.
Since we assume that the UAV is already roughly oriented towards the door, we restrict this search to $\pm \SI{60}{\degree}$ around the current heading.
The detection reliability is maximized when the door's surface normal aligns with the radial direction of the obstacle scan.
Thus, we remove all detections exhibiting a large deviation from this alignment.
Furthermore, this helps to remove false positives caused by jumps in the distance scan due to occlusions.
Finally, all candidates are checked for sufficient width and clearance behind the opening.
If multiple suitable detections remain, we choose the one whose normal aligns best with the current UAV heading.
\Reffig{fig:door_detection} shows an example of the door detection.

In a next step, a pre-traversal pose is computed:
We project the end points of the door opening into 3D and horizontally offset the midpoint along the outward normal.
The target yaw is chosen to align with the direction of the inward normal.
To compensate for drift or inaccuracies in the obstacle measurements, we only steer the UAV a small distance towards the pre-traversal pose and update the target on receiving the next obstacle scan.

When the UAV reaches a stable pre-traversal pose, we fix the estimated doorway position and steer the UAV along the doorway normal toward the post-traversal pose placed behind the door opening.
Lateral obstacle avoidance is disabled while traversing the door.

\section{Evaluation}
\label{sec:evaluation}

\subsubsection{Control Latency}
In a first experiment, we evaluate the latency of our pipeline.
We measure the time from sending a motion command at the GCS until the movement is visualized to the operator.
Additionally, we record the delay until the UAV starts moving using a motion capture system.
On average, it took \SI{496}{\milli\second} until a motion started and another \SI{344}{\milli\second} for the updated VIO estimate to be visualized at the GCS.
The round-trip time of the WiFi connection between GCS and Android RCU was only \SI{4.5}{\milli\second}.
The total latency of \SI{839}{\milli\second} emphasizes the necessity for our \textit{carrot mode} as manual control with raw velocity commands becomes infeasible at this latency.

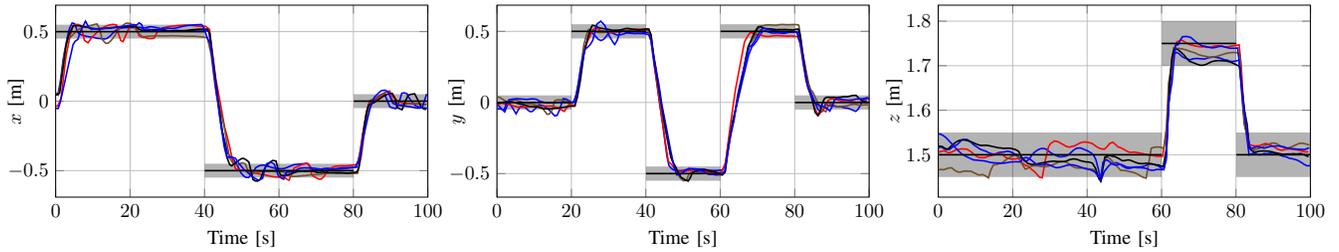
\begin{figure*}[t!]
	\centering
	\resizebox{\linewidth}{!}{%
\begin{tikzpicture}%
\begin{groupplot}[
group style={
group name=my plots,
group size=3 by 1,
vertical sep=1.5em,
horizontal sep=3.5em,
},
width=0.95\columnwidth,
height=5cm,
xlabel={Time [s]},
ylabel style={rotate=-90},
grid=both,
xmin=0, xmax=100,
cycle list name=color,
every axis plot/.append style={no markers, thick}
]

\nextgroupplot[ylabel={$x$ [m]}, ylabel style={rotate=90}, ylabel shift=-1.5em,
 xlabel={Time [s]}]
\addplot table[x index=0, y index=1, col sep=space, each nth point=10] {control_data/odom_1};
\addplot table[x index=0, y index=1, col sep=space, each nth point=10] {control_data/odom_2};
\addplot table[x index=0, y index=1, col sep=space, each nth point=10] {control_data/odom_3};
\addplot table[x index=0, y index=1, col sep=space, each nth point=10] {control_data/odom_4};
\addplot table[x index=0, y index=1, col sep=space, each nth point=10] {control_data/odom_5};
\addplot[jump mark left, black] table[x index=0, y index=1, col sep=space] {control_data/task};
\addplot[name path=target_upper, jump mark left, draw=none] table[x index=0, y index=1, col sep=space] {control_data/task_lower};
\addplot[name path=target_lower, jump mark left, draw=none] table[x index=0, y index=1, col sep=space] {control_data/task_upper};
\addplot[fill=black, opacity=0.3] fill between[of=target_upper and target_lower];

\nextgroupplot[ylabel={$y$ [m]}, ylabel style={rotate=90}, ylabel shift=-1.5em,
 xlabel={Time [s]}]
\addplot table[x index=0, y index=2, col sep=space, each nth point=10] {control_data/odom_1};
\addplot table[x index=0, y index=2, col sep=space, each nth point=10] {control_data/odom_2};
\addplot table[x index=0, y index=2, col sep=space, each nth point=10] {control_data/odom_3};
\addplot table[x index=0, y index=2, col sep=space, each nth point=10] {control_data/odom_4};
\addplot table[x index=0, y index=2, col sep=space, each nth point=10] {control_data/odom_5};
\addplot[jump mark left, black] table[x index=0, y index=2, col sep=space] {control_data/task};
\addplot[name path=target_upper, jump mark left, draw=none] table[x index=0, y index=2, col sep=space] {control_data/task_lower};
\addplot[name path=target_lower, jump mark left, draw=none] table[x index=0, y index=2, col sep=space] {control_data/task_upper};
\addplot[fill=black, opacity=0.3] fill between[of=target_upper and target_lower];

\nextgroupplot[ylabel={$z$ [m]}, ylabel style={rotate=90}, ylabel shift=-0.5em,
 xlabel={Time [s]}]
\addplot table[x index=0, y index=3, col sep=space, each nth point=10] {control_data/odom_1};
\addplot table[x index=0, y index=3, col sep=space, each nth point=10] {control_data/odom_2};
\addplot table[x index=0, y index=3, col sep=space, each nth point=10] {control_data/odom_3};
\addplot table[x index=0, y index=3, col sep=space, each nth point=10] {control_data/odom_4};
\addplot table[x index=0, y index=3, col sep=space, each nth point=10] {control_data/odom_5};
\addplot[jump mark left, black] table[x index=0, y index=3, col sep=space] {control_data/task};
\addplot[name path=target_upper, jump mark left, draw=none] table[x index=0, y index=3, col sep=space] {control_data/task_lower};
\addplot[name path=target_lower, jump mark left, draw=none] table[x index=0, y index=3, col sep=space] {control_data/task_upper};
\addplot[fill=black, opacity=0.3] fill between[of=target_upper and target_lower];
\end{groupplot}
\end{tikzpicture} %
}
	\vspace{-0.5cm}
	\caption{Evaluating position control accuracy. Horizontal black lines depict five target positions with the gray-shaded area indicating to the target threshold. The colored lines show the trajectories of five different runs as estimated by the VIO.}
	\label{fig:control}
\end{figure*}

\subsubsection{Flight Towards a Waypoint}
DJI lets us control the UAV via velocity commands with a resolution of \SI{0.1}{\meter\per\second} at a frequency of \SI{10}{\hertz}.
To analyze how accurately we reach a given target, we autonomously fly towards five different poses using a target threshold of \SI{5}{\centi\meter} per axis.
The resulting trajectories (estimated from VIO) for five repeated flights are shown in \reffig{fig:control}.
We successfully reached all targets in all tries.
Occasionally, the UAV drifted out of the target zone but our control directly steered it back.
We tracked the distance between the VIO and target poses, from first entering the zone until the next target was sent.
On average, we achieve an accuracy of \SI{4.7}{\centi\meter} with a std. dev. of \SI{1.2}{\centi\meter}.

\begin{figure}
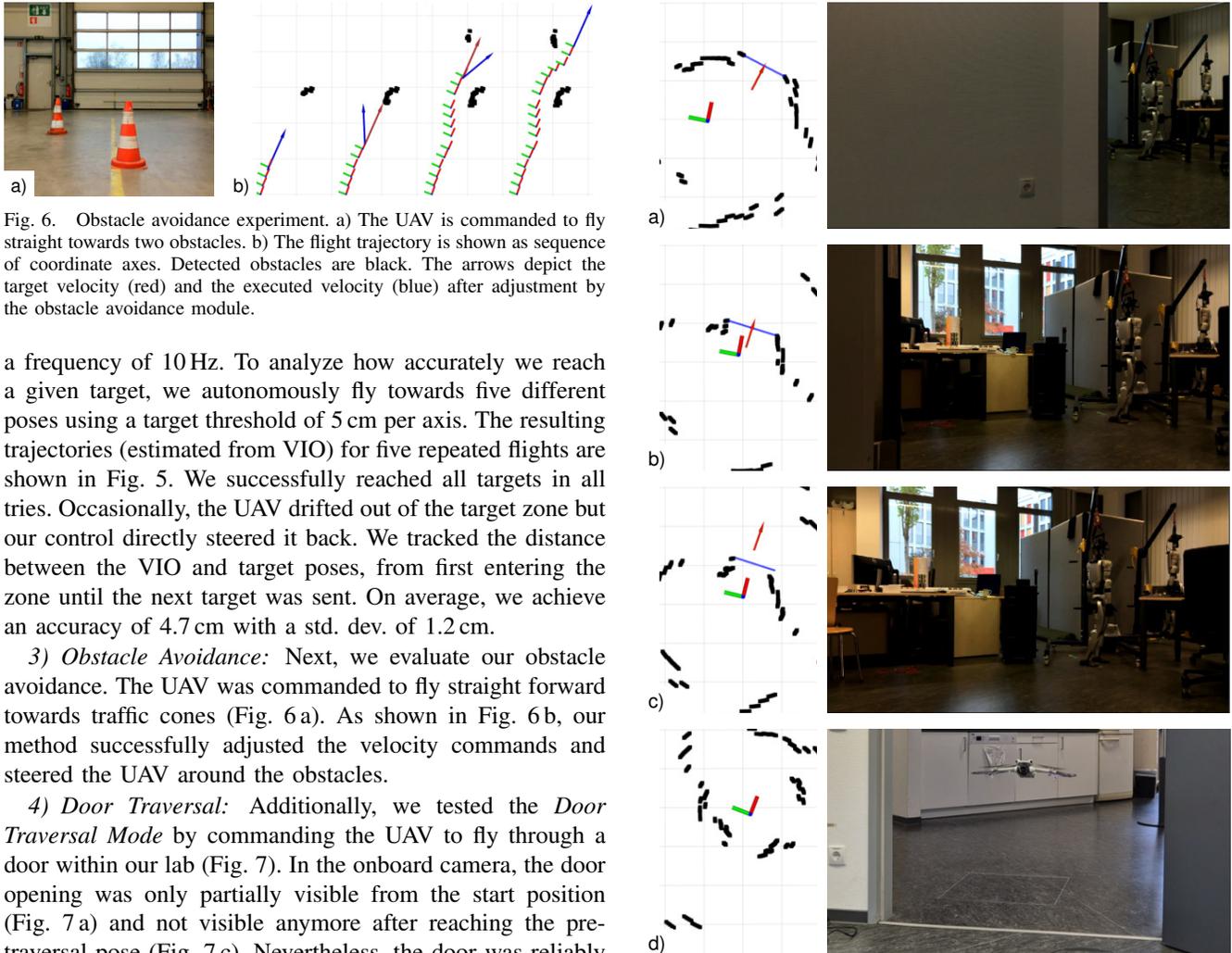

	\centering
	\resizebox{\linewidth}{!}{%
\begin{tikzpicture}

\node[inner sep=0,anchor=north west] (image0) at (0,0)
{\includegraphics[trim=135 0 140 30,clip,height=2cm]{figures/avoidance_onboard_1.png}};

\node[inner sep=0,anchor=north west,xshift=0.4cm] (image1) at (image0.north east)
{\includegraphics[trim=50 0 30 0,clip,height=2.0cm]{figures/avoidance_map_1.png}};

\node[inner sep=0,anchor=north west,xshift=0cm] (image2) at (image1.north east)
{\includegraphics[trim=50 0 30 0,clip,height=2.0cm]{figures/avoidance_map_2.png}};

\node[inner sep=0,anchor=north west,xshift=0cm] (image3) at (image2.north east)
{\includegraphics[trim=50 0 30 0,clip,height=2.0cm]{figures/avoidance_map_3.png}};

\node[inner sep=0,anchor=north west,xshift=0cm] (image4) at (image3.north east)
{\includegraphics[trim=50 0 30 0,clip,height=2.0cm]{figures/avoidance_map_4.png}};

\node[scale=0.6, anchor=south west,xshift=-0.01cm,yshift=-0.1cm, rectangle, fill=white, align=center, font=\small\sffamily] (n_0) at (image0.south west) {a)};

\node[scale=0.6, anchor=south east,xshift=-1.4cm,yshift=-0.1cm, rectangle, fill=white, align=center, font=\small\sffamily] (n_0) at (image1.south east) {b)};

\end{tikzpicture} %
}
	\vspace{-0.5cm}
	\caption{Obstacle avoidance experiment. a) The UAV is commanded to fly straight towards two obstacles. b) The flight trajectory is shown as sequence of coordinate axes. Detected obstacles are black. The arrows depict the target velocity (red) and the executed velocity (blue) after adjustment by the obstacle avoidance module.}
	\label{fig:avoidance}
\end{figure}

\subsubsection{Obstacle Avoidance}
Next, we evaluate our obstacle avoidance.
The UAV was commanded to fly straight forward towards traffic cones (\reffig{fig:avoidance}\,a).
As shown in \reffig{fig:avoidance}\,b, our method successfully adjusted the velocity commands and steered the UAV around the obstacles.

\begin{figure}[!t]
	\centering
	\resizebox{\linewidth}{!}{%
\begin{tikzpicture}

\node[inner sep=0,anchor=north west] (image0) at (0,0)
{\includegraphics[trim=210 90 90 40,clip,height=2.2cm]{figures/door_onboard_1.png}};

\node[inner sep=0,anchor=north west,xshift=0.1cm] (image1) at (image0.north east)
{\includegraphics[height=2.2cm]{figures/door_cam_1.png}};

\node[inner sep=0,anchor=north west,yshift=-0.15cm] (image2) at (image0.south west)
{\includegraphics[trim=210 90 90 40,clip,height=2.2cm]{figures/door_onboard_2.png}};

\node[inner sep=0,anchor=north west,xshift=0.1cm] (image3) at (image2.north east)
{\includegraphics[height=2.2cm]{figures/door_cam_2.png}};

\node[inner sep=0,anchor=north west,yshift=-0.15cm] (image4) at (image2.south west)
{\includegraphics[trim=210 90 90 40,clip,height=2.2cm]{figures/door_onboard_3.png}};

\node[inner sep=0,anchor=north west,xshift=0.1cm] (image5) at (image4.north east)
{\includegraphics[height=2.2cm]{figures/door_cam_3.png}};

\node[inner sep=0,anchor=north west,yshift=-0.15cm] (image6) at (image4.south west)
{\includegraphics[trim=210 90 90 40,clip,height=2.2cm]{figures/door_onboard_4.png}};

\node[inner sep=0,anchor=north west,xshift=0.1cm] (image7) at (image6.north east)
{\includegraphics[trim=0 5 0 0,clip,height=2.2cm]{figures/door_external.jpg}};

\node[scale=0.6, anchor=south west,xshift=-0.3cm,yshift=-0.1cm, rectangle, fill=white, align=center, font=\small\sffamily] (n_0) at (image0.south west) {a)};

\node[scale=0.6, anchor=south west,xshift=-0.3cm,yshift=-0.1cm, rectangle, fill=white, align=center, font=\small\sffamily] (n_0) at (image2.south west) {b)};

\node[scale=0.6, anchor=south west,xshift=-0.3cm,yshift=-0.1cm, rectangle, fill=white, align=center, font=\small\sffamily] (n_0) at (image4.south west) {c)};

\node[scale=0.6, anchor=south west,xshift=-0.3cm,yshift=-0.1cm, rectangle, fill=white, align=center, font=\small\sffamily] (n_0) at (image6.south west) {d)};

\end{tikzpicture}
}
	\vspace{-0.5cm}
	\caption{Door traversal experiment. Obstacles scans including the current UAV pose (axes), door detections (blue line) and current target pose (red arrow) are depicted on the left. a)-c) show corresponding onboard footage, while d) shows an external view of the UAV passing the door. }
	\label{fig:door_traversal}
\end{figure}

\subsubsection{Door Traversal}
Additionally, we tested the \textit{Door Traversal Mode} by commanding the UAV to fly through a door within our lab (\reffig{fig:door_traversal}).
In the onboard camera, the door opening was only partially visible from the start position (\reffig{fig:door_traversal}\,a) and not visible anymore after reaching the pre-traversal pose (\reffig{fig:door_traversal}\,c).
Nevertheless, the door was reliably detected within the obstacle scan during the whole flight.
The estimated door width ranged from \SI{0.78}{\meter} to \SI{1.38}{\meter}, with an average of \SI{1.10}{\meter} and a std. dev. of \SI{0.12}{\meter}, thus slightly underestimating the actual width of \SI{1.25}{\meter} most of the time.
The detected door orientation ranged from  \SI{0.2}{\degree} to \SI{30}{\degree}, with average \SI{6.7}{\degree} and std. dev. \SI{6.4}{\degree}.
The inaccuracies in the estimation of the door pose were successfully handled by continuously updating the pre-traversal pose to the most recent detections.
\Reffig{fig:door_traversal}\,d shows that the UAV passed through the center of the door opening with sufficient safety clearance to the door frame.

\subsubsection{3D Mapping}
\begin{figure}[t]
	\centering
	\resizebox{0.8\linewidth}{!}{\includegraphics{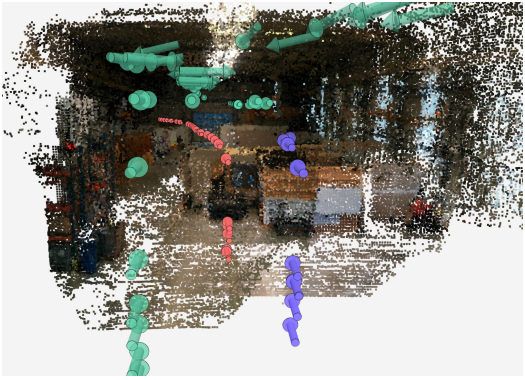}}
	\caption{Reconstructed colored 3D map from images of three UAVs. Colored arrows represent the UAV flight trajectories.}
	\label{fig:reconstruction}
\end{figure}

Finally, we evaluate the global environment mapping.
\Reffig{fig:reconstruction} shows an example of a reconstructed 3D point cloud from three different UAVs (DJI Mini\,3\,Pro, Mini\,4\,Pro, and Mavic\,3T) flying simultaneously in the DRZ Living Lab.
Additionally, we tested our system in an outdoor scenario.
A densely aggregated LiDAR point cloud~\cite{QuenzelB:IROS21} provides a reference for the reconstruction when either GNSS or odometry poses are used.
For the comparison, we compute the point-to-point distance to the LiDAR map. 
To cope with unrepresented areas within the LiDAR map, we threshold distances above \SI{0.5}{\metre} and \SI{1}{\metre}. 
The mean distance is \SI{0.081}{\metre} at threshold \SI{0.5}{\metre} for both sets of poses. 
However, we obtain \SI{0.135}{\metre} for the odometry and \SI{0.150}{\metre} using GNSS at threshold \SI{1}{\metre}.
Moreover, the odometry poses are more consistent as emphasized by the 33\% higher count of triangulated points whereas the percentage of points above the threshold is similar.

\section{Conclusion \& Future Work}
We presented a remote autonomy pipeline for flight in GNSS-denied environments using small consumer-grade UAVs.
We developed an Android app for the remote controller that steers the UAV towards target poses while successfully avoiding obstacles.
The ground control station fuses the perception of multiple UAVs into a global 3D environment model, while offering an intuitive interface for configuration and supervision of multiple UAVs by a single operator.
As a next step, we plan to conduct a user study in a simulated rescue operation to evaluate the usability of our approach under realistic conditions by real first responders.

\if\WithAuthorInfo1
\section*{Acknowledgement}
\small{We would like to thank Subham Agrawal, Nils Dengler and Maren Bennewitz from the Humanoid Robots Lab at the
University of Bonn for their help with the Motion Capture System.

This work has been supported by the German Federal Ministry of Research, Technology and Space (BMFTR) in the project ``Kompetenzzentrum: Etablierung des Deutschen Rettungsrobotik-Zentrums (E-DRZ)'', grant 13N16477.}
\fi

\printbibliography

\end{document}